\documentclass{IEEEtran4PSCC}
\usepackage{ifpdf}
\usepackage{epstopdf}

\ifCLASSINFOpdf
  \usepackage[pdftex]{graphicx}
  \graphicspath{{../pdf/}{../jpeg/}}
  \DeclareGraphicsExtensions{.pdf,.jpeg,.png}
\else
  \usepackage[dvips]{graphicx}
  \graphicspath{{../eps/}}
  \DeclareGraphicsExtensions{.eps}
\fi
\usepackage[cmex10]{amsmath}
\usepackage{amsfonts}
\begin{document}
\title{Ensemble Methods of Classification for Power
Systems Security Assessment}

%% To specify the authors when (number of affiliations <= 2)
\author{
\IEEEauthorblockN{Alexei Zhukov, Victor Kurbatsky \\ Nikita Tomin, Denis Sidorov, Daniil Panasetsky }
\IEEEauthorblockA{Energy Systems Institute \\
Russian Academy of Sciences\\
Irkutsk, Russia\\
\{kurbatsky, dsidorov, tomin\}@isem.sei.irk.ru}
\and
\IEEEauthorblockN{Aoife Foley}
\IEEEauthorblockA{School of Mechanical and Aerospace Engineering \\
Queens University Belfast\\
Belfast, UK\\
a.foley@qub.ac.uk}
}

%% To specify the authors when (number of affiliations > 2)
% \author{\IEEEauthorblockN{Author n.1\IEEEauthorrefmark{1},
% Author n.2\IEEEauthorrefmark{2},
% Author n.3\IEEEauthorrefmark{3}, 
% Author n.4\IEEEauthorrefmark{3} and
% Author n.5\IEEEauthorrefmark{4}}
% \IEEEauthorblockA{\IEEEauthorrefmark{1} Department Name of Organization A\\
% Name of the organization A,
% Address A\\ Emails if wanted}
% \IEEEauthorblockA{\IEEEauthorrefmark{2} Department Name of Organization B\\
% Name of the organization B,
% Address B\\ Emails if wanted}
% \IEEEauthorblockA{\IEEEauthorrefmark{3} Department Name of Organization C\\
% Name of the organization C,
% Address C\\ Emails if wanted}
% \IEEEauthorblockA{\IEEEauthorrefmark{4}Department Name of Organization D\\
% Name of the organization D,
% Address D\\ Emails if wanted}
% }

\maketitle

\begin{abstract}
One of the most promising approaches for complex technical systems analysis employs ensemble methods of classification. Ensemble methods enable to build a reliable decision rules for feature space classification in the presence of many possible states of the system. In this paper, novel techniques based on decision trees are used for evaluation of the reliability of the regime of electric power systems. We proposed hybrid approach based on random forests models and boosting models. 
Such techniques can be applied to predict the interaction of increasing renewable power, strage devices and swiching of smart loads from intelligent domestic appliances, storage heaters and air-conditioning units and electric vehicles with grid for enhanced decision making.
The ensemble classification methods were tested on the modified 118-bus IEEE power system showing that proposed technique can be employed to examine whether the power system is secured under steady-state operating conditions. 
\end{abstract}


% Use this to place sponsorships
\thanksto{
This work is funded by the RSF project ``Development of an intelligent system for preventing large-scale emergencies in power systems'' under grant  No. 14-19-00054.}

\section{Introduction}
Assessment of security of bulk electric power systems is one of the pressing problems in the modern power engineering. The trends towards liberalization and the need to increase electricity transmission due to growing loads and generation expansion make existing power companies operate  their electrical networks in critical conditions, close to their admissible security limits [1].  In such conditions the unforeseen excess disturbances, weak connections, hidden defects of the relay protection system and automated devices, human factors as well as a great amount of other factors can cause a drop in the system security or even the development of catastrophic accidents.

For the time being there is a wide spectrum of approaches and tools for the assessment of security. All the variety of the methods can be divided into:
\begin{itemize}
	\item Traditional approaches based on a detailed modeling of potential disturbances in electric power systems and numerical calculations of nonlinear capacity equations [3, 4];
  \item Intelligent approaches which involve the artificial intelligence algorithms learning on a limited set of power system states, such as artificial neural networks, support vector machine, decision trees, etc. [1, 2, 5, 19].
\end{itemize}

An analysis of methods for the assessment of security and voltage stability of electric power system shows that the existing traditional approaches cannot be effectively applied in the online and real time conditions because of their computation complexity.  For example load flow calculation for the assessment of the aftermath of a system component fault, which underlies the classical approaches to the assessment of security in electric power systems and does not seem to be fully implemented because of complex modeling of corresponding protections, that trip the overloaded line or load feeder in case of inadmissible voltage level. Moreover, currently to meet the demand for electricity as well to ensure quality and reliability of electricity supply systems, the distributed generation is connected as a local energy source.

Although, many of the developed approaches on the basis of intelligent models are more adapted to real changes in the power system topology and system conditions, they still do not have sufficient sensitivity to these changes, not always offer the possibility of predicting voltage stability losses, and do not allow us to estimate the probability of identifying a potentially dangerous state.

At the transmission level (i.e. bulk power), phasor measurement units (PMU's) have been introduced to improve grid reliability. A PMU is a calculated real time phasor measurement synchronised to an absolute reference unit provided by a Global Positioning System (GPS). These PMU's are used to assess grid (e.g. MVARs, kV, frequency changes etc.)  conditions because they synchronise power quality in real time by comparing phase angle measurements. To date PMU are only deployed for Wide Area Measurement (WAM) [20, 21]. One of the issues of applying and using the large amounts of PMU datasets for rapid decision making. The decision making and onus is usually still with the expertise of the grid operators. However, as the number of market participants, renewable power sources, storage devices and smart loads increase in the power system both at the transmission (and distribution) level the decision making will become ever more complex. Hence this research employs
 the ensemble methods on the basis of decision trees. The calculations involved the following modifications of random forest models (Extremely Randomized Trees, Oblique Random Forests) and boosting models (Stochastic Gradient Boosting, AdaBoost). The effectiveness of their application is confirmed by a great number of calculations on the basis of power system test scheme. The suggested ensemble methods of classification are implemented in the free software environment R intended for calculations with an open-source code.

\section{Problem Statement}
Security is an ability of electric power system to withstand sudden disturbances without unforeseen effects  on the electricity consumers. It is provided by control capabilities of power systems. In the operational practices the required level of security can be achieved by both the preventive control actions (before a disturbance) and the emergency control actions (after disturbance). Control in the pre-emergency conditions is mainly a responsibility of the operational dispatching control. At the same time there can be situations where the speed of power system control by the dispatching personnel appears to be insufficient to avoid dangerous situations. The complexity of a problem here lies in the fact that most of dangerous (pre-emergency) states of electric power system which lead to  large-scale blackouts are unique and there is no single “algorithm” (for solving) to effectively reveal  such conditions  as the time. The problem gets complicated by the fact that the security limit of electric power system constantly changes, therefore fast methods for real time security monitoring are required to analyze the current level of security and accurately trace the limit and detect the most vulnerable regions along it. 

The key idea of the pre-emergency control concept is the fact that the voltage instability following the emergency disturbance which accompanies many system emergencies does not develop as fast as the dynamic instability of the electric power system [6]. Thus, when the phase of slow emergency development comes, the balance between generation and consumption is maintained for a long time   and this makes it possible to detect potentially dangerous states, which appear after the disturbance, and generate respective preventive control actions [2].

\section{Ensemble Algorithms of Classification and the Problem of Power Security Assessment}
A great many studies show that the effective solution to this problem can be found on the basis of machine learning  methods which normally include artificial neural networks, decision trees, ensemble (committee) models, etc. This is related to their capabilities of fast detection of the images, patterns (i.e. typical  samples), learning/generalization and, which is important, high speed of identifying the instability  boundaries.

One of the advanced approaches to the analysis of complex technical systems is ensemble methods of classification. They make it possible to form reliable decision rules of classification  for a set of potential system states. In this approach the key idea is to build a universal classifier of power system states which is capable of tracing dangerous pre-emergency conditions and predicting emergency situations on the basis of certain system security indices. In this case the detection of dangerous operation patterns is  not effective without considering probable disturbance/faults, whose calculation leads to a considerable increase in the computational complexity and a potential decrease in the accuracy for basic algorithms. This leads to the need of finding a way to improve the accuracy of the classifier of power system states. One of such methods is the creation of ensembles of the classification models and their training. 

One of the first most general theory of algorithmic ensembles was proposed in the algebraic approach by Y.I. Zhuravlev [7]. According to this theory the composition of $N$ basic algorithms $h_t=C(a_t (x))$, $t=1, \dots, N$ is taken to mean a superposition of algorithmic operators $a_t: \, X\rightarrow {\mathbb R}$, of a correction operation ${\mathbf F}: {\mathbb R}^N\rightarrow {\mathbb R}$ and decision rule ${\mathcal C}: {\mathbb R}\rightarrow Y$ such as $H(x)={\mathcal C}\left({\mathbf F}(a_1 (x), \dots , a_N (x))\right)$, where $x \in X$, $X$ is a space of objects, $Y$ is a set of answers, and ${\mathbb R}$ is a space of estimates. 

Later Valiant and Kearns \cite{IEEEhowto:kearns} were the first to pose the question about whether or not a weak learning algorithm can be strengthened to an arbitrary accurate learning algorithm. This process was called boosting. Schapire [9] developed the first provable polynomial-time boosting algorithm. It was intended for the conversion of weak models into strong ones by constructing an ensemble of classifiers. The main idea of the boosting algorithm is step-by-step enhancement of the algorithm ensemble. One of the popular implementations of this idea is Schapire’s  AdaBoost algorithm which involves ensemble of decision trees [10].

Another approach to the classification and regression problems using the ensembles was suggested by Leo Breiman [11]. This approach is an extension of the bagging idea. According to this idea, a collective decision can be obtained by using an elementary committee method which classifies an object according to a decision of most of the algorithms. Unlike the boosting method bagging is based on parallel learning of base classifiers.
One of the progressive bagging-based approaches is the method called Random Forest [11, 12]. Later there appeared the most effective modifications of both Random forests and boosting algorithms such  as Extremely Randomized Trees, Oblique Random Forests and Stochastic Gradient Boosting  [13].

In the researches devoted to the security assessment there are many approaches oriented to the construction of models on the basis of decision trees [1, 2, 5,13]. These models use both off-line (periodically updated) and on-line methods. Single trees are easily interpretable, yet do not always allow us to obtain the required accuracy when approximating complex target relationships. Therefore, it is considered reasonable to use compositions [14].

\section{Calculation of a Power System Security on the Basis of Ensemble Models} 
Figure 1 presents a general scheme of the suggested approach for the estimation of power system security. The primary principle of the approach lies in the mathematical model learning on the basis of the ensemble method of classification to automatically make a sufficiently accurate  assessment of the power system conditions according to the criterion secure/insecure on the basis of significant  classification attributes of a power system state, for example active and reactive power flows, bus voltage, etc.  A great amount of such attributes are obtained on the basis of randomly generated data sample consisting of a set of really possible states of electric power system [2]. 

\begin{figure}[!ht]
\centering
\includegraphics[width=3.6in]{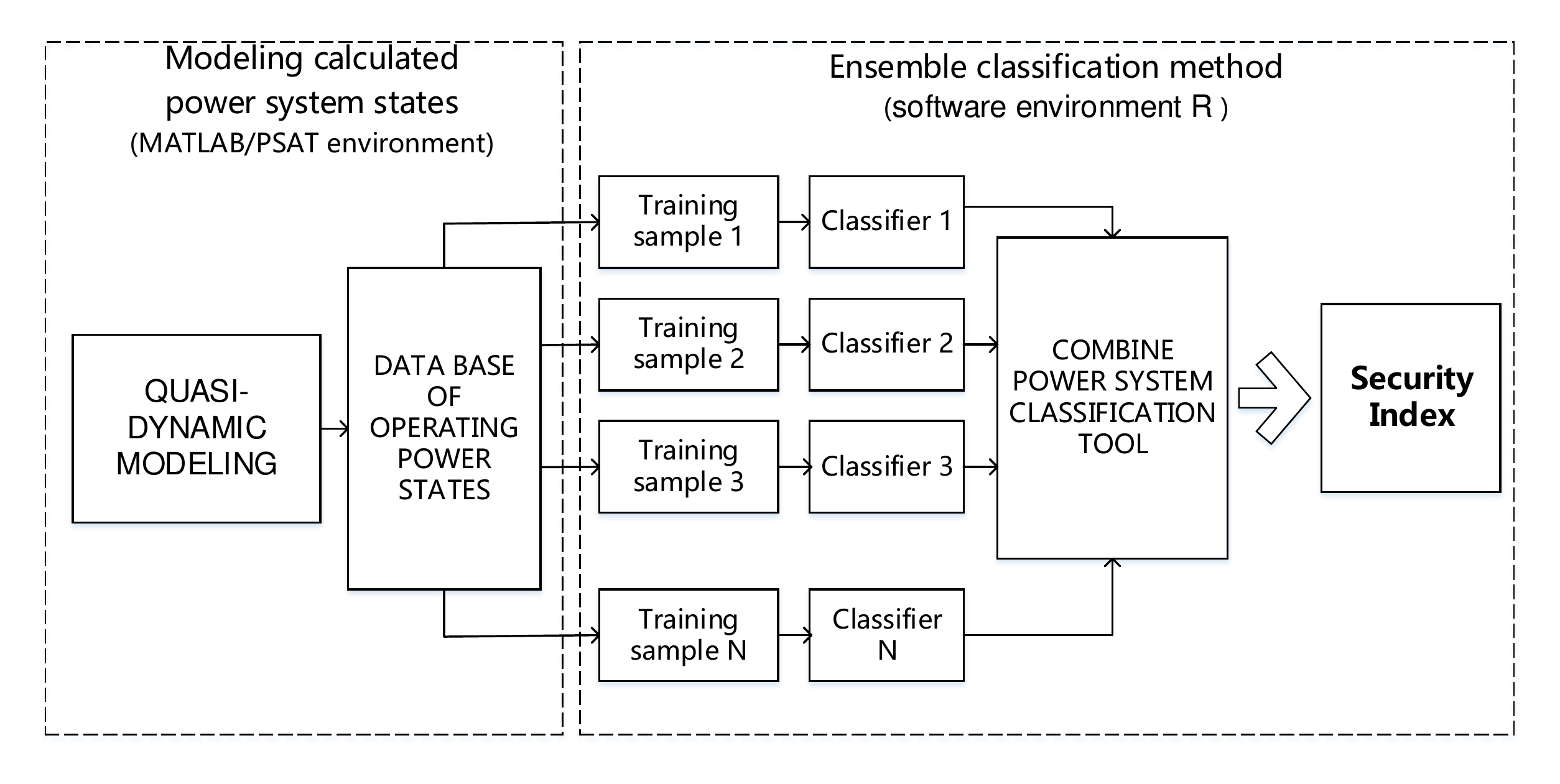}
% where an .eps filename suffix will be assumed under Latex, 
% and a .pdf suffix will be assumed for pdfLatex; or what has been declared
\DeclareGraphicsExtensions.
\caption{A general scheme of the assessment of potential power system security, using compositional models.}
\label{fig1}
\end{figure}

Depending on the ensemble method applied each decision rule will be trained by its subsampling according to the bagging and boosting principles. The final decision on the classification of any power system state is made within the generalized classifier according to different principles of simple majority voting, weighted voting or by choosing the most competent decision rule. 

In the paper a list of potential power system states for the model learning  is formed using quasi-dynamic modeling with a special  program in the MATLAB environment (Fig.2). The load model was represented by static characteristics depending on voltage. When critical values of voltage are achieved the load is automatically transferred to shunts. The method of a proportional increase in load at all nodes of the test scheme was optimized for the security analysis in such a way that the initial condition for each emergency disturbance is a stable condition  closest to  it, from those calculated. Thus, at each stage of an increase in the test scheme load the emergency events (primary disturbances) are randomly modeled by the $N-1$ reliability principle. As a result, the database including a set of various pre-emergency and emergency states of the test scheme is built. 

\begin{figure}[h]
\centering
\includegraphics[width=3.5in]{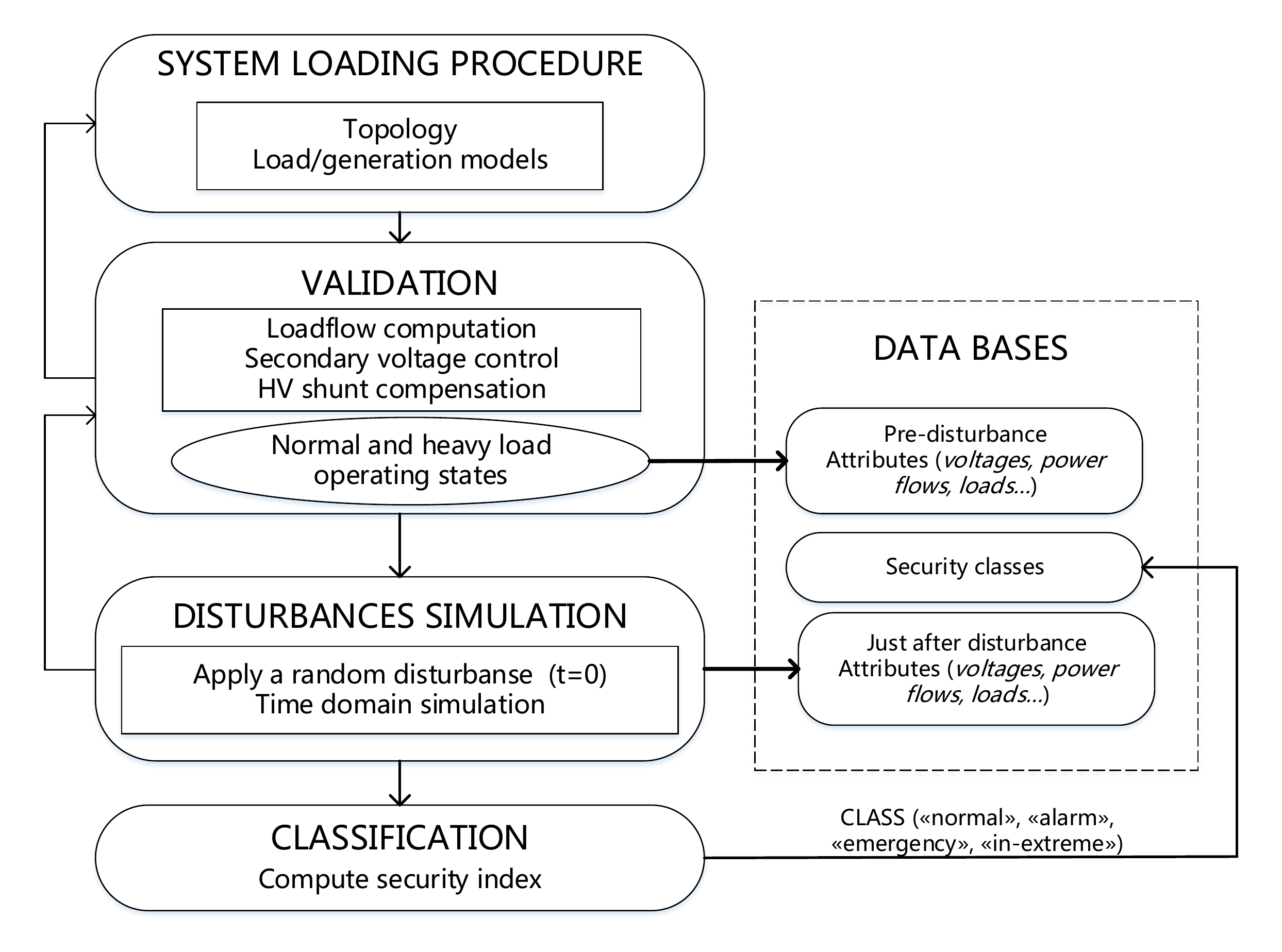}
% where an .eps filename suffix will be assumed under Latex, 
% and a .pdf suffix will be assumed for pdfLatex; or what has been declared
\DeclareGraphicsExtensions.
\caption{A scheme of modeling calculated states of electric power system in the problem of security monitoring and assessment on the basis of quasi-dynamic modeling in the MATLAB/PSAT environment.}
\label{fig2}
\end{figure}

In order to obtain a problem book each of such states is assigned a security index (here readers may refer to [14])  which is calculated by the following expression:
\begin{equation}
\label{eqn:43} 
SI = {w_1\cdot \sum\nolimits_{i=1}^{n_L} LOI_i + w_2\cdot \sum\nolimits_{i=1}^{n_B} VDI_i \over {n_L+n_B}},
\end{equation}
where  $w_1$ and $w_2$ are the weighting factors of system security; $LOI$ is  line overload index; $VDI$ is index of voltage deviation at nodes; $n_L$ and $n_B$ represent the number of lines and buses respectively. 

Thus, the security index is determined by calculating the VDI and LOI, which are obtained using  the following expressions:

\begin{equation}
%\label{eqn:example}
LOI_{km}=\begin{cases}
{{S_{km} - S_{lim}} \over {S_{km}}} \cdot {100}, &\text{if $S_{km} > S_{lim}$;}\\
0, &\text{if $S_{km} < S_{lim}$.}
\end{cases}
\end{equation}

\begin{equation}
%\label{eqn:example}
VDI_{k}=\begin{cases}
{{|U_{k}^{min}| - |U_{k}| } \over |U_{k}^{min}|} \cdot {100}, &\text{if $|U_{k}| < |U_{k}^{min}|$;}\\
0, &\text{if $|U_{k}^{min}| \leq|U_{k}| \leq |U_{k}^{max}|$;} \\
{{|U_{k}|-|U_{k}^{max}|} \over |U_{k}^{max}|} \cdot {100}, &\text{if $|U_{k}| > |U_{k}^{max}|,$}\\
\end{cases}
\end{equation}
where  $S_{km}$ and $S_{lim}$ represent the MVA flow and MVA limit of branch k-m correspondingly, $|U_{k}|$, $|U_{k}^{\max}|$ and $|U_{k}^{\max}|$ are the minimum voltage limit, maximum voltage limit and bus voltage magnitude of k-bus respectively.

Evaluating the security index as given by (\ref{eqn:43}), each pattern is labeled as belonging to one of the four classes as shown in Table \ref{tab:ClassLabelsForPowerSecurityAnalisys}.

\begin{table}[ht]
	\centering
	\begin{tabular} {|c|c|}
		\hline
		Security Index & Class Category/Power State \\ 
		\hline
		$SI=0$  & Normal state \\ 
		\hline
		$0 < SI \leq 5\%$ & Alarm state \\ 
		\hline
		$5\% < SI\leq 15\%$ & Emergency 1 state \\ 
		\hline
		 $SI > 15\%$ & Emergency 2 state \\ 
		\hline
		\end{tabular}
		\label{tab:ClassLabelsForPowerSecurityAnalisys}
	\caption{Class labels for power security analysis.} 
\end{table}

To give an idea of the criteria without going into details, the system states are presented.

\begin{itemize}
\item Normal state implies that all parameters of the power system are maintained within specified normal operation limits.
\item Alarm state that some of the system parameters exceed the specified normal limits (for example, bus voltage can exceed ± 5\%, but remain within ± 10\%). Depending on the operation rules, actions can take place to bring the system to the normal state. 
\item Emergency 1 state implies the system is still intact; however, some system constraints are violated. The system can be restored to the normal state (or at least to the alarm state), if the suitable corrective actions are taken. 
\item Emergency 2 state implies that the current situation cannot be corrected and will lead to major emergency. Control actions, like load shedding or controlled system separation are used for saving as much of the system as possible from a widespread blackout.
\end{itemize}

In the case where the values of the indices exceed the specified limits on security and the high probability of emergency situations that correspond to these values, respective preventive or emergency control measures can be formed.

\section{Case study}
The feasibility of the approach in a proof-of-concept has been demonstrated on the IEEE 118 power system consisting of more than 118 buses, 54 generators, and 186
transmission lines.  An open-source environment R [16] with  \textbf{caret} package
%\cite{caret2015manual} 
 [17] is used as a computing environment 
for proposed models design and testing. 

\subsection{Data base generation}

Operating conditions are all generated using the Powertrain System Analysis Toolkit (PSAT) [18]. The database generation and data conversion are conducted in MATLAB. The overall data base generation procedure, whose aim was to provide a representative sample of possible power system states, combining various prefault operating states and disturbances, is illustrated in Fig. 2.

A data base representative of the potential power system states was obtained by generating, firstly, a sample of various prefault situations, and applying to each state the random disturbances to produce the corresponding stability scenarios. To obtain the data base composed of 6877 states, each of the prefault normal or heavy load states was combined with the possible disturbances. These have been simulated with a variable step MATLAB/PSAT quasi-dynamic simulation program, which computed the attribute values and allowed us to classify based on the security index the scenarios as normal, alarm, emergency 1 and emergency 2. The 490 initial candidate attributes such as active and reactive power flow, voltages used to characterize the power system states. They represent essentially power system quantities which may be available from the SCADA system  or the Phasor Data Concentrator system in the power system state.

\subsection{Estimating Performance For Classification}
In current paper we need to use proper performance measurement metrics for classification problems. We used the following metrics:
\begin{itemize}
	\item The overall accuracy of a model indicates how well the model predicts the actual data.
	\item the Kappa statistic takes into account the expected error rate:
		\begin{equation}
%\label{eqn:example}
k = {{ O - E } \over {1 - E}}
\end{equation}
where O is the observed accuracy and E is the expected accuracy
under chance agreement. 
\end{itemize}

\subsection{Overview of obtained results}
Ensemble and single trees methods have been built for classifying the power system states, for various candidate attributes and four different security classifications. Tab.  II %\ref{fig:accuracyComparison} 
shows comparison of accuracy achieved by the state-of-the-art  classification tree learning algorithms. Namerly, the following classifiers where tested:
J48 decision tree, conventional Breiman's non-parametric decision tree learning technique CART, bagged CART (BCART),  Random Forest, Extra Trees (ET) and 
 Stochastic Gradient Descent (SGB) method. The ensemble and single tree methods were trained on 6877 samples dataset and tested on 1715 samples. Confusion matrix is shown in Tab. I. Each column of the matrix represents the instances in a predicted class while each row represents the instances in an actual class

\begin{table}[ht]
	\centering
	\begin{tabular}{|l|l|l|l|l|l|l|}
		\hline
		%\multirow{Metrics}
{Metrics} & \multicolumn{6}{c|}{Ensemble Methods} 
 \\ \cline{2-7} 
		& J48 & CART & BCART & RF & ET & SGB \\ 
		\hline
		Accuracy \% & 99.83 & 99.07 & 99.88 & 100.00 & 100.00 & 100.00 \\ 
		\hline
		Kappa \% & 99.72 & 98.52  & 99.81 & 100.00 & 100.00 & 100.00 \\ 
		\hline
	\end{tabular}
	\label{fig:accuracyComparison}
	\caption{Classification accuracy comparison.} 
\end{table}

	\begin{table}[ht]
				\centering
				\begin{tabular}{|r|r|r|r|r|r|}
					\hline
					& alarm & emergency1 & emergency2 & normal & class error \\ 
					\hline
					alarm & 947.00 & 1.00 & 0.00 & 0.00 & 0.00 \\ 
					emerg1 & 4.00 & 279.00 & 5.00 & 0.00 & 0.03 \\ 
					emerg2 & 0.00 & 4.00 & 252.00 & 0.00 & 0.02 \\ 
					normal & 2.00 & 0.00 & 0.00 & 225.00 & 0.01 \\ 
					\hline
				\end{tabular}
\label{cmatrix}
				\caption{Confusion Matrix.}	
			\end{table}

\begin{figure}[h]
		\centering
		\includegraphics[scale=0.30]{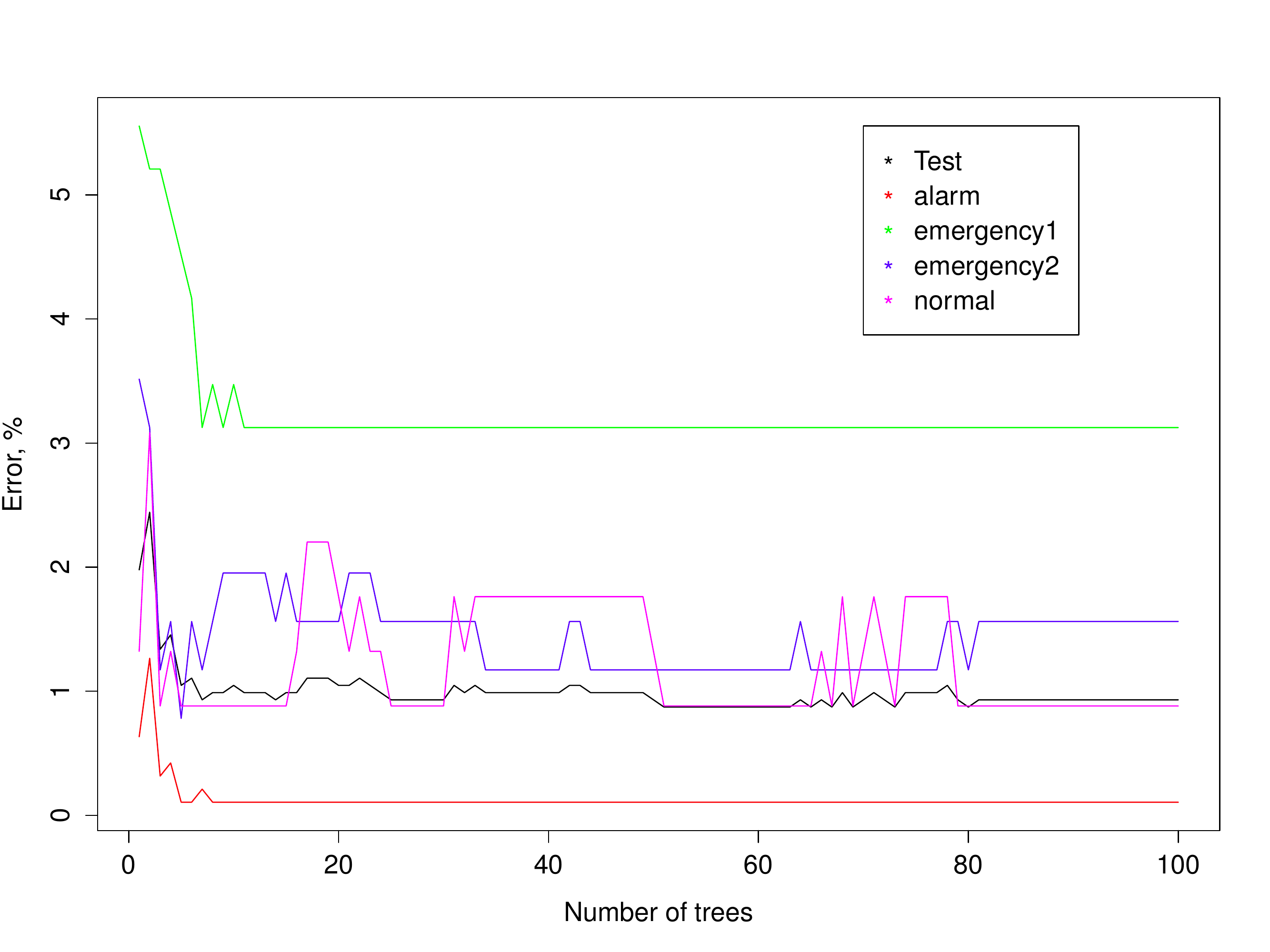}		  	
		\caption{Testing error with respect to the number of trees.}
		\label{fig:ieee118-rf-testcurve}
	\end{figure}

Fig. %\ref{fig:ieee118-rf-traincurve} and 
\ref{fig:ieee118-rf-testcurve}   shows %training and
 testing errors with respect to the number of trees for normal, alarm, emergency1 and emergency2 cases. As footnote its to be noted that all obtained accuracy values are close. But some of the models enjoy additional useful properties. For example Extra Trees needs less memory comparing with classical Random Forest but comparable with Stohastical gradient boosting.

\begin{figure}[ht!]
		\centering
		\includegraphics[scale=0.26]{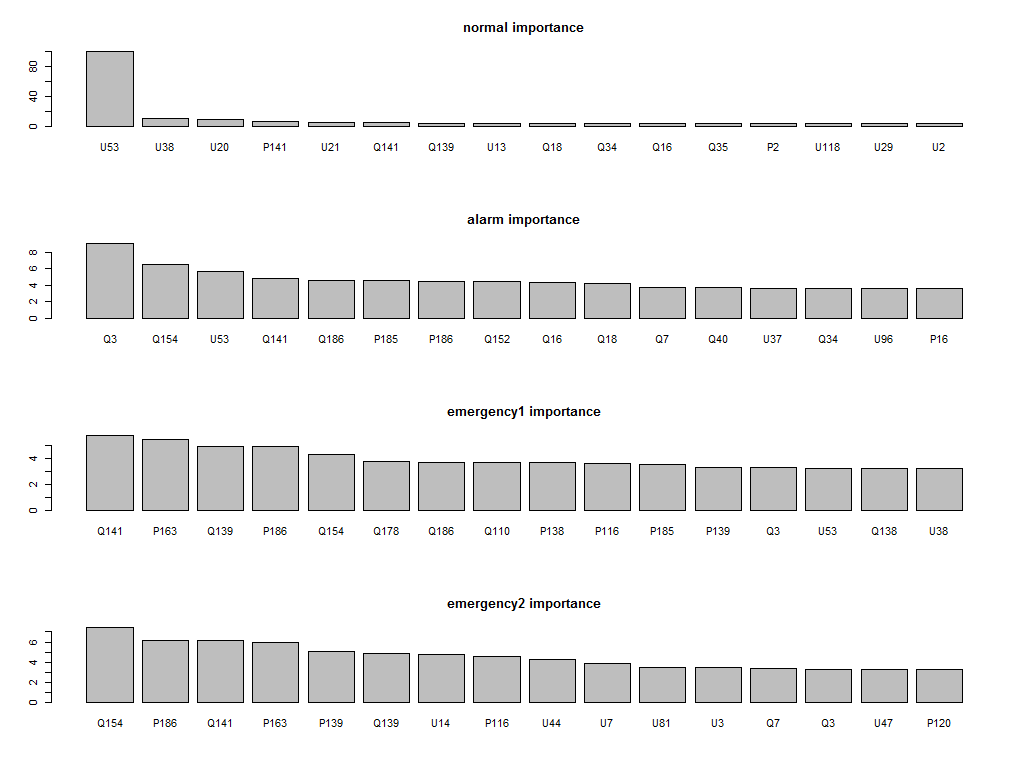}		  	
		\caption{Variable importance for all classes obtained by computing of mean gini index decrease.}
		\label{fig:RF_model_importance_by_class}
	\end{figure}

Fig. \ref{fig:RF_model_importance_by_class} shows variable importance for all classes obtained by computing of mean gini index decrease. The classification trees select voltages under normal states as the most important attributes for security monitoring and assessment. It may be “explained” by the fact that the voltage sag observed in the power system state reflects proportional increase in load, when the static characteristics of the load model depend on voltage. Under alarm and emergency states the power and reactive flow attributes were selected in preference to voltages. A possible explanation lies in the fact that this security criterion is more “preventive like”. 

We also demonstrated the feasibility of dealing with incomplete and distorted data. Taking into consideration SCADA malfunctions, the corrupted patterns were used to train ensemble classification trees. The results showed that the test error rate did not changed even if 50\% of gaps (Tab. IV).

\begin{table}[ht!]
				\centering
				\begin{tabular}{|l|l|l|}
					 \hline
					 \% of gaps & time in sec. & test error, \% \\
					 \hline
					 10 & 0.0123 & 0.93 \\
					 30 & 0.0411 & 0.93 \\
					 50 & 0.0514 & 0.93 \\
					 \hline
				\end{tabular}
				\caption{Filling the gaps in data} 
			\end{table}

%Out-of-bag (OOB) error estimate is also included in Fig. \ref{fig:ieee118-rf-traincurve}. 
 
%	\begin{figure}[h]
%		\centering
%		\includegraphics[scale=0.25]{ieee118-rf-traincurve.png}		  	
%		\caption{Training error with respect to the number of trees.}
%		\label{fig:ieee118-rf-traincurve}
%	\end{figure}

%{~}\bigskip
%\begin{center}
%\Huge{Case Study for PSCC 2015}
%\end{center}
%
%\section{Case Study}
%The proposed ensemble methods are tested on an standard IEEE scheme (on Fig.\ref{fig:model_photo_system}) consisting of more than 118 buses, 54 generators, and 186
%transmission lines.
%
%	\begin{figure}[t]
%		\centering
%		\includegraphics[scale=0.13]{IEEE118BusSystemA}		  	
%		\caption{IEEE 118-buses scheme}
%		\label{fig:model_photo_system}
%	\end{figure}
	
%All proposed models models are trained and tested using na open-source environment R \cite{R2014manual} with \textbf{caret} package\cite{caret2015manual}.

%\section{Ensemble methods}

\section{Conclusion}
The ensemble classification methods were tested on the modified IEEE 118 power system showing that proposed technique can be employed to examine whether the power system is secured under steady-state operating conditions. The experimental studies showed that the ensemble methods can identify key system parameters as security indicators with high accuracy and, if required, the obtained security tree-based model can produce an alarm for triggering emergency control system. The next stage of this work will involve taking real power system data and modelling multiply decision making with many grid participants.

\end{document}